\documentclass{article}

\usepackage[nonatbib,preprint]{neurips_2020}

\usepackage[utf8]{inputenc} 
\usepackage[T1]{fontenc}    
\usepackage{hyperref}       
\usepackage{url}            
\usepackage{booktabs}       
\usepackage{amsfonts}       
\usepackage{nicefrac}       
\usepackage{microtype}      
\usepackage{amsmath}
\usepackage[linesnumbered,ruled,vlined]{algorithm2e}

\usepackage{graphicx}
\usepackage{bm}
\usepackage{multirow}
\usepackage{multicol}

\usepackage{subcaption}

\title{Multi-branch Attentive Transformer\thanks{This work is conducted at Microsoft Research Asia.}}

\author{%
$^1$Yang Fan, $^2$Shufang Xie, $^2$Yingce Xia, $^2$Lijun Wu, $^2$Tao Qin, $^1$Xiang{-}Yang Li, $^2$Tie{-}Yan Liu\\
$^1$ University of Science and Technology of China\quad$^2$ Microsoft Research Asia\\
  $^1$\texttt{fyabc@mail.ustc.edu.cn,\;xiangyangli@ustc.edu.cn}\\\quad$^2$\texttt{\{shufxi,yingce.xia,lijun.wu,taoqin,tyliu\}@microsoft.com} \\
}

\newcommand{\concat}{\texttt{concat}}

\newcommand{\multiattn}[1]{\texttt{attn}_{#1}}
\newcommand{\maattn}{\texttt{mAttn}}
\newcommand{\attn}{\texttt{attn}}
\newcommand{\ffn}{\texttt{FFN}}
\newcommand{\hffn}{\texttt{hFFN}}

\begin{document}

\maketitle

\begin{abstract}
While the multi-branch architecture is one of the key ingredients to the success of computer vision tasks, it has not been well investigated in natural language processing, especially sequence learning tasks. In this work, we propose a simple yet effective variant of Transformer~\cite{vaswani2017attention} called multi-branch attentive Transformer (briefly, MAT), where the attention layer is the average of multiple branches and each branch is an independent multi-head attention layer. We leverage two training techniques to regularize the training: drop-branch, which randomly drops individual branches during training, and proximal initialization, which uses a pre-trained Transformer model to initialize multiple branches. Experiments on machine translation, code generation and natural language understanding demonstrate that such a simple variant of Transformer brings significant improvements. Our code is available at  \url{https://github.com/HA-Transformer}.
\end{abstract}

\section{Introduction}\label{sec:intro}
The multi-branch architecture of neural networks, where each block consists of more than one parallel components, is one of the key ingredients to the success of deep neural models and has been well studied in computer vision.
Typical structures include the inception architectures~\cite{szegedy2015going,szegedy2016rethinking,szegedy2017inception}, ResNet~\cite{he2016deep}, ResNeXt~\cite{xie2017aggregated}, DenseNet~\cite{huang2017densely}, and the network architectures discovered by neural architecture search algorithms~\cite{pham2018efficient,liu2019darts}. Models for sequence learning, a typical natural language processing problem, also benefit from multi-branch architectures. Bi-directional LSTM (BiLSTM) models can be regarded as a two-branch architecture, where a left-to-right LSTM and a right-to-left LSTM are incorporated into one model. BiLSTM has been applied in neural machine translation (briefly, NMT)~\cite{zhou2016deep,wu2016google} and pre-training~\cite{peters2018elmo}. Transformer~\cite{vaswani2017attention}, the state-of-the-art model for sequence learning, also leverages multi-branch architecture in its multi-head attention layers. 

Although multi-branch architecture plays an important role in Transformer, it has not been well studied and explored. Specifically, using hybrid structures with both averaging and concatenation operations, which has been widely used in image classification tasks~\cite{xie2017aggregated,szegedy2016rethinking}, are missing in current literature for sequence learning. This motivates us to explore along this direction. We propose a simple yet effective variant of Transformer, which treats a multi-head attention layer as a branch, duplicates such an attention branch for multiple times, and averages the outputs of those branches. Since both the concatenation operation (for the multiple attention heads) and the averaging operation (for the multiple attention branches) are used in our model, we call our model {\em multi-branch attentive Transformer} (briefly, MAT) and such an attention layer with both the concatenation and adding operations as the multi-branch attention layer.


Due to the increased structure complexity, it is challenging to directly train MAT. Thus, we leverage two techniques for MAT training. (1) {\em Drop branch}: During training, each branch should be independently dropped so as to avoid possible co-adaption between those branches. Note that similar to Dropout, during inference, all branches are used. (2) {\em Proximal initialization}: We initialize MAT with the corresponding parameters trained on a standard single-branch Transformer.

Our contributions are summarized as follows: 

\noindent(1) We propose a simple yet effective variant of Transformer, multi-branch attentive Transformer (MAT). We leverage two techniques, drop branch and proximal initialization, to train this new variant. 

\noindent(2) We conduct experiments on three sequence learning tasks: neural machine translation, code generation and natural language understanding. On these tasks, MAT significantly outperforms the standard Transformer baselines, demonstrating the effectiveness of our method. 

\noindent(3) We explore another variant which introduces multiple branches into feed-forward layers. We find that such a modification slightly hurts the performance. 


\section{Background}\label{sec:background}

\subsection{Introduction to Transformer}
A Transformer model consists of an encoder and a decoder. Both the encoder and the decoder are stacks of blocks. Each block is mainly made up of two types of layers: the multi-head attention layer and the feed-forward layer (briefly, FFN). We will mathematically describe them.

Let $\concat(\cdots)$ denote the concatenation operation, where all inputs are combined into a larger matrix along the last dimension. Let $\multiattn{M}$ and $\attn$ denote a multi-head attention layer with $M$ heads ($M\in\mathbb{Z}_+$) and a standard attention layer. A standard  attention layer~\cite{bahdanau2015neural,vaswani2017attention} takes three elements as inputs, including query $Q$, key $K$ and value $V$, whose sizes are $T_q\times d$, $T\times d$ and $T\times d$ ($T_q, T, d$ are integers). $\attn$ is defined as follows: 
\begin{equation}
\attn(Q,K,V)=\texttt{softmax}\Big(\frac{QK^\top}{\sqrt{d}}\Big)V,
\end{equation}
where \texttt{softmax} is the softmax operation. A multi-head attention layer aggregates multiple attention layers in the concatenation way:
\begin{equation}
\begin{aligned}
& \multiattn{M}(Q,K,V)=\concat(H_1,H_2,\cdots,H_M);
\;H_i=\attn(QW^i_Q,KW^i_K,VW^i_V),\;i\in[M],
\end{aligned}
\label{eq:multi_head_attention}
\end{equation}
where $[M]$ denotes the set $\{1,2,\cdots,M\}$. In Eqn.\eqref{eq:multi_head_attention}, the $W$'s are the parameters to be learned. Each $W$ is of dimension $T\times (d/M)$. The output of $\multiattn{M}$ is the same size as $Q$, i.e., $T_q\times d$.

In standard Transformer, an FFN layer, denoted by $\ffn$, is implemented as follows:
\begin{equation}
\ffn(x)=\max(xW_1 + b_1, 0)W_2 + b_2,
\end{equation}
where $x$ is a $1\times d$-dimension input, $\max$ is an element-wise operator, $W_1$ and $W_2$ are $d\times d_h$ and $d_h\times d$ matrices, $b_1$ and $b_2$ are $d_h$ and $d$ dimension vectors. Usually, $d_h>d$.


In the encoder of a Transformer, each block consists of a self-attention layer, implemented as a $\multiattn{M}$ where the query $Q$, key $K$ and value $V$ are the outputs of previous layer, and an FFN layer. In the decoder side, an additional multi-head attention is inserted between the self-attention layer and FFN layer, which is known as the encoder-decoder attention: $Q$ is the output of the previous block, $K$ and $V$ are the outputs of the last block in the encoder. 

\subsection{Multi-branch architectures for sequence learning}
As shown in Section~\ref{sec:intro}, multi-branch architectures have been well investigated in image processing. In comparison, the corresponding work for sequence learning is limited. The bidirectional LSTM, a two-branch architecture, has been applied in machine translation~\cite{zhou2016deep,wu2016google} and pre-training techniques like ELMo~\cite{peters2018elmo}. \cite{song2018double} proposed two use both convolutional neural networks and Transformer in the encoder and decoder. Similar idea is further expanded in \cite{zhao2019muse}. A common practice of the previous work is that they focus on which network components (convolution, attention, etc.) should be used in a multi-branch architecture. In this work, we do not want to introduce additional operations into Transformer but focus on how to boost Transformer with its own components, especially, where and how to apply the multi-branch topology, and figure out several useful techniques to train multi-branch architectures for sequence learning. Such aspects are missing in previous literature.

\section{Multi-branch attentive Transformer}\label{sec:alg}
In this section, we first introduce the structure of multi-branch attentive Transformer (MAT) in Section~\ref{sec:network_arch}, and then we introduce the drop branch technique in Section~\ref{sec:drop_branch}. The proximal initialization is described in Section~\ref{sec:prox_init}. 

\subsection{Network architecture}\label{sec:network_arch}
The network architecture of our proposed MAT adopts the backbone of standard Transformer, except that all multi-head attention layers (including both self-attention layers and encoder-decoder attention layers) are replaced with the {\em multi-branch attention layers}.

Let $\maattn_{N_a,M}(Q,K,V)$ denote a multi-branch attention layer, where $N_a$ represents the number of branches and each branch is a multi-head attention layer with $M$ heads $\multiattn{M}$. $Q$, $K$ and $V$ denote query, key and value, which are defined in Section~\ref{sec:background}. Mathematically, $\maattn_{N_a,M}(Q,K,V)$ works as follows:
\begin{equation}
\maattn_{N_a,M}(Q,K,V)=Q + \frac{1}{N_a}\sum_{i=1}^{N_a}\multiattn{M}(Q,K,V;\theta_i),
\label{eq:multi-branch_attention_layer_infer}
\end{equation}
where $\theta_i = \{W^i_Q,W^i_K,W^i_V\}_{i=1}^{M}$ (see Eqn.\eqref{eq:multi_head_attention}) is the collection of all parameters of the $i$-th multi-head attention layer. 

\subsection{Drop branch technique}\label{sec:drop_branch}
As mentioned above, multiple branches in multi-branch attention layers have the same structure. To avoid the co-adaptation among them, we leverage the drop branch technique. The inspiration comes from the dropout~\cite{srivastava2014dropout} and drop path technique~\cite{larsson2017fractalnet}, where some branches are randomly dropped during training.

Let $\maattn_{N_a,M}(Q,K,V;\rho)$ denote a multi-branch attention layer with drop branch rate $\rho\in[0,1]$, which is an extension of that in Section~\ref{sec:network_arch}. Equipped with drop branch technique, the $i$-th branch of the multi-branch attention layer, denoted as $\beta^M_i(Q,K,V;\rho,\theta_i)$, works as follows:
\begin{equation}
\beta^M_i(Q,K,V;\rho,\theta_i)=\frac{\mathbb{I}\{U_i \ge \rho\}}{1-\rho}\multiattn{M}(Q,K,V;\theta_i),
\end{equation}
where $U_i$ is uniformly sampled from $[0,1]$ (briefly, $U_i\sim\texttt{unif}[0,1]$); $\mathbb{I}$ is the indicator function.
During training, we may set $\rho\ge0$, where any branch might be skipped with probability $\rho$. During inference, we must set $\rho=0$, where all branches are leveraged with equal weights $1$. The multi-branch attention layer $\maattn_{N_a,M}(Q,K,V;\rho)$ works as follows:
\begin{equation}
\maattn_{N_a,M}(Q,K,V;\rho)=Q + \frac{1}{N_a}\sum_{i=1}^{N_a}\beta^M_i(Q,K,V;\rho,\theta_i).
\label{eq:multi-branch_attention_layer}
\end{equation}
Note when $\rho=0$, Eqn.\eqref{eq:multi-branch_attention_layer} degenerates to Eqn.\eqref{eq:multi-branch_attention_layer_infer}.

During training, it is possible that all branches in multi-branch attention layer are dropped. The residual connection ensures that even if all branches are dropped, the output from the previous layer can be fed to top layers through the identical mapping. That is, the network is never blocked. When $N_a=1$ and $\rho=0$, a multi-branch attention layer degenerates to a vanilla multi-head attention layer. When $N_a=1$ and $\rho>0$, it is the standard Transformer with randomly dropped attention layers during training, which is similar to the drop layer~\cite{fan2020reducing}. (See Section~\ref{sec:hyper_explore} for more discussions.)

\begin{figure}[!htbp]
\centering
\includegraphics[width=0.95\linewidth]{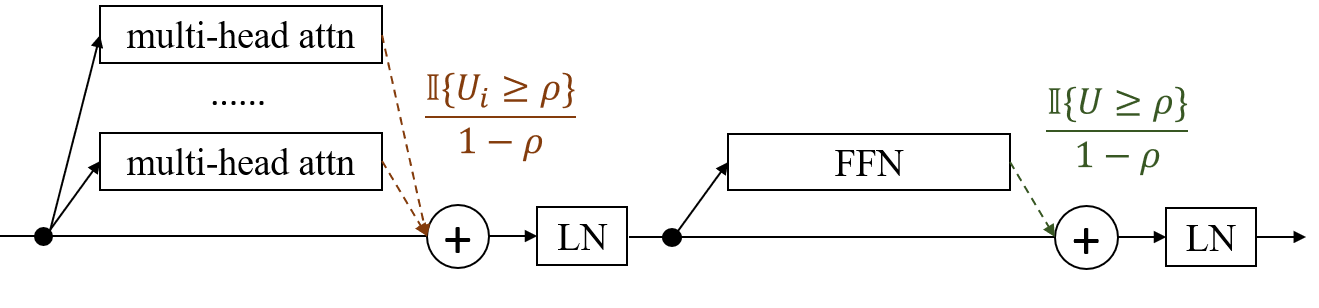}
\vspace{-3mm}
\caption{Architecture of a block in the encoder of MAT. ``LN'' refers to layer normalization.}
\label{fig:arch_multi_branch}
\vspace{-3mm}
\end{figure}

We also apply the drop branch technique to FFN layers. That is, the revised FFN layer is defined as 
\begin{equation}
x +\frac{\mathbb{I}\{U\ge\rho\}}{1-\rho}\ffn(x),\;U\sim\texttt{unif}[0,1]
\end{equation}
where $x$ is the output of the previous layer. An illustration of multi-branch attentive Transformer is in Figure~\ref{fig:arch_multi_branch}. Each block in the encoder consists of a multi-branch attentive self-attention layer and an FFN layer. Each block in the decoder consists of another multi-branch attentive encoder-decoder attention layer. Layer normalization~\cite{ba2016layer} remains unchanged.

\subsection{Proximal initialization}\label{sec:prox_init}
MAT can be optimized like the standard version by first randomly initializing all parameters and  training until convergence. However, the multi-branch attention layer increases the training complexity of MAT, since there are multiple branches to be optimized.

Recently, proximal algorithms~\cite{neal2013proximal} have been widely used in pretraining-and-finetuning framework to regularize training~\cite{jiang2019smart}. Then main idea of proximal algorithms is to balance the trade-off between minimizing the objective function and minimizing the weight distance with a pre-defined weight. Inspired by those algorithms, we design a two-stage warm-start training strategy:

\noindent(1) Train a standard Transformer with embedding dimension $d$, FFN dimension $d_h$.

\noindent(2) Duplicate both the self-attention layers and encoder-decoder attention layers for $N_a$ times to initialize MAT. Train this new model until convergence.
 
We empirically find that the proximal initialization scheme can boost the performance than that obtained by training from scratch.

\section{Application to neural machine translation}
In this section, we conduct two groups of experiments: one with relatively small scale data, including IWSLT'14 German$\to$English, IWSLT'14 Spanish$\leftrightarrow$English and IWSLT'17 French$\leftrightarrow$English translation tasks; the other with larger training corpus, WMT'14 English$\to$German translation and WMT'19 German$\to$French translation. We briefly denote English, German, Spanish and French as En, De, Es and Fr respectively.   

\subsection{Settings}
\noindent{\em Data preprocessing}:
Our implementation of NMT experiments is based on fairseq\footnote{\url{https://github.com/pytorch/fairseq}}.
For IWSLT'14 De$\leftrightarrow$En, we follow~\cite{edunov2018classical} to get and preprocess the training, validation and test data\footnote{The URLs of scripts we used in this paper are summarized in Appendix C.}, including lowercasing all words, tokenization and applying BPE~\cite{sennrich2016bpe}. For the other two IWSLT tasks, we do not lowercase the words and keep them case-sensitive. We apply tokenization and BPE as those used in preprocessing IWSLT De$\leftrightarrow$En. Training data sizes for IWSLT De$\to$En, Es$\leftrightarrow$En and Fr$\leftrightarrow$En are $160k$, $183k$ and $236k$ respectively. We choose IWSLT'13 Es$\leftrightarrow$En, IWSLT'16 Fr$\leftrightarrow$En for validation purpose, and choose IWSLT'14 Es$\leftrightarrow$En and IWSLT'17 Fr$\leftrightarrow$En as test sets. The numbers of BPE merge operation for the three tasks are all $10k$. The source and target data are merged to get the BPE table.

For WMT'14 English$\to$German translation, we follow \cite{ott2018scaling} to preprocess the data , and eventually obtain $4.5M$ training data. We use newstest 2013 as the validation set, and choose newstest 2014 as the test set. The number of BPE merge operation is $32k$. The preprocess steps for WMT'19 De$\to$Fr are the same as those for WMT'14 En$\to$De, and we eventually get $9M$ training data in total. We concatenate newstest2008 to newstest 2014 together as the validation set and use newstest 2019 as the test set.

\noindent{\em Model configuration and training strategy}: For the three IWSLT tasks, we choose the default setting provided by fairseq official code\footnote{\url{https://github.com/pytorch/fairseq/blob/master/fairseq/models/transformer.py}} as the baseline with embedding dimension $d=512$, hidden dimension $d_h=1024$ and number of heads $M=4$. For WMT'14 En$\to$De, we mainly follow the big transformer setting, where the above three numbers are $1024$, $4096$ and $16$ respectively. The dropout rates are $0.3$. For the more detailed parameters like $\rho$ and $N_a$, we will introduce the details in corresponding subsections. We use the Adam~\cite{kingma2015adam} optimizer with initial learning rate $5\times10^{-4}$, $\beta_1=0.9$, $\beta_2=0.98$ and the \texttt{inverse\_sqrt} learning rate scheduler~\cite{vaswani2017attention} to control training. Each model is trained until convergence. The source embedding, target embedding and output embedding of each task are shared. The batch size is 4096 for both IWSLT and WMT tasks. For IWSLT tasks, we train on single P40 GPU; for WMT tasks, we train on eight P40 GPUs.

\noindent{\em Evaluation} We evaluate the translation quality by BLEU scores. For IWSLT'14 De$\to$En and WMT'14 En$\to$De, following the common practice, we use \texttt{multi-bleu.perl}. For other tasks, we choose \texttt{sacreBLEU}.


\subsection{Exploring hyper-parameters of MAT}\label{sec:hyper_explore}
Due to resource limitation, we first explore hyper-parameters and proximal initialization on IWSLT'14 De$\to$En dataset to get some empirical results, then transfer them to larger datasets.

We try different combination of $N_a$, $d$ and $d_h$. We ensure the number of total parameters not exceeding $36.7M$, the size of the default model for IWSLT'14 De$\to$En. All results are reported in Table~\ref{tab:iwslt_deen_multi-branch_attn}, where the network architecture is described by a four-element tuple, with each position representing the number of branches $N_a$, embedding dimension $d$, hidden dimension $d_h$ and model sizes.

\begin{table}[!htbp]
\vspace{-1mm}
\centering
\small
\caption{Results on IWSLT'14 De$\to$En with different architectures. The left and right subtables are experiments of standard transformer and MAT respectively. From left to right in each subtable, the columns represent the network architecture, number of parameters, BLEU scores with $\rho$ ranging from $0.0$ to $0.3$.}
\begin{tabular}{p{2.44cm}p{.62cm}p{.56cm}p{.56cm}p{.56cm}|p{2.44cm}p{.56cm}p{.56cm}p{.56cm}p{.56cm}}
\toprule
$N_a/d/d_h/\text{Param}$ & $0.0$ & $0.1$ & $0.2$ & $0.3$ & $N_a/d/d_h/\text{Param}$ & $0.0$ & $0.1$ & $0.2$ & $0.3$ \\
\midrule
\multicolumn{5}{c|}{Standard Transformer + Drop Branch} & \multicolumn{5}{c}{MAT + Drop Branch} \\ 
\midrule
$1/512/1024/36.7M$ &    $34.95^\triangle$ & $3.94$ &    $22.43$ &   $0.78$ & $2/256/1024/18.4M$ &    $34.42$ & $35.19$ & $35.52$ & $35.11$ \\
$1/256/1024/13.7M$ &	$35.04$ &	$35.39$ &	$34.53$ &	$29.34$ & $2/256/2048/24.7M$ &    $34.51$ & $35.46$ & $35.59$ & $35.33$ \\
$1/256/2048/20.0M$ &	$34.66$ &	$35.45$ &   $32.75$ &	$1.37$ & $3/256/1024/23.1M$ &    $34.01$ & $34.92$ &	$35.39$ &	$35.44$ \\
$1/256/3072/26.3M$ & 	$34.41$ &	$35.37$ &	$34.64$ &	$25.05$ & $3/256/2048/29.4M$ &    $33.98$ & $35.03$ &	$35.40$ &	$\bm{35.70}$ \\
 & & & & & $4/256/1024/27.9M$ & 	$33.77$ &	$34.84$ & $35.08$ &	$35.14$ \\
 & & & & & $4/256/2048/34.2M$ & 	$33.79$ &	$34.81$ &	$35.08$ &	$35.46$ \\
\bottomrule
\end{tabular}
\label{tab:iwslt_deen_multi-branch_attn}
\vspace{-1mm}
\end{table}

The baselines correspond to the architectures with $N_a=1$ and $\rho=0$. The most widely adopted baseline (marked with $^\triangle$) is $34.95$ and the model size is $36.7$M. Reducing $d$ to $256$ results in slightly BLEU score $35.04$. We have the following  observations:

\noindent(1) Using multi-branch attention layers with more than one branches can boost the translation BLEU scores, with a proper $\rho$. When setting $d_h=1024$ and increasing $N_a$ from $2$ to $3$, the BLEU scores are $35.52$ (with $\rho=0.2$) and $35.44$ (with $\rho=0.3$), outperforming the standard baseline $34.95$, as well as the BLEU score obtained from architecture $1/256/1024$. However, it is not always good to enlarge $N_a$. The BLEU score of MAT with $N_a=4$ is $35.14$, which is a minor improvement over the baseline.

Multi-branch attention layers also benefits from larger hidden dimension $d_f$. When setting $d_f=2048$, we obtain consistent improvement compared to $d_f=1024$ with the same $N_a$. Among all architectures, we find that the model $3/256/2048$ achieves the best BLEU score, which is $35.70$ (we also evaluate the validation perplexity of all architectures, and the model $3/256/2048$ still get the lowest perplexity $4.71$). Compared with the corresponding Transformer $1/256/2048$ with $\rho=0$, we get $1.04$ improvement. We also enlarge $\rho$ to $0.4$ and $0.5$, but the BLEU scores drop. Results are shown in Appendix A.

\noindent(2) The drop branch is important technique for training multi-branch attention layers. Take the network architecture $3/256/1024$ as an example: If we do not use drop branch technique, i.e., $\rho=0$, we can only get $34.01$ BLEU score, which is $0.94$ point below the baseline. As we increase $\rho$ from $0$ to $0.3$, the BLEU scores become better and better, demonstrating the effectiveness of drop branch. 

\noindent(3) When setting $N_a=1$ and $\rho_B>0$, the architecture still benefits from the drop branch technique. For architecture $1/256/1024$, the vanilla baseline is $35.04$. As we increase $\rho$ to $0.1$, we can obtain $0.35$ point improvement. Wider networks with $d_f$ also benefits from this technique, which shows that drop branch is generally a useful trick for Transformer. This is consistent with the discoveries in~\cite{fan2020reducing}. With $\rho=0.1$, enlarging the hidden dimension to $2048$ and $3072$ lead to $35.45$ and $35.37$ BLEU scores, corresponding to $0.79$ and $0.96$ score improvements compared with that without using drop branch. But we found that when $N_a=1$, drop branch might lead to unstable results. For example, when $\rho=0.1$, $1/512/1024$ cannot lead to a reasonable result. We re-run the experiments with five random seeds but always fail. In comparison, MAT can always obtain reasonable results with drop branch.

We also try to turn off the drop branch at the FFN layers. In this case, we found that by ranging $\rho$ from $0$ to $0.3$, architecture $3/256/2048$ can achieve $34.63$, $34.78$ and $34.98$ BLEU scores, which are worse than those obtained by using drop branch at all layers. This shows that the drop branch at every layers is important for our proposed model.

\subsection{Exploring proximal initialization}\label{sec:proximal_init}
In this section, we explore the effect of proximal initialization, i.e., warm start from an existing standard Transformer model. The results are reported in Table~\ref{tab:result_iwslt_deen}. Generally, compared with the results without proximal initialization, the models can achieve more than $0.5$ BLEU score improvement. Specifically, with $N_a=3$ and $d_h=2048$, we can achieve a $36.22$ BLEU score, setting a state-of-the-art record on this task. We also evaluate the validation perplexity of all architectures in Table~\ref{tab:result_iwslt_deen}, and the model with $N_a=3$ and $d_h=2048$ still get the lowest perplexity $4.56$.

\begin{table}[!htbp]
\vspace{-2mm}
\centering
\caption{Results of using proximal initialization. Columns from left to right represent the network architecture, BLEU scores with drop branch ratio $\rho$ from $0.0$ to $0.3$, and the increment compared to the best results without proximal initialization.}
\begin{tabular}{cccccccc}
\toprule
$(N_a,d_h)$&  $0.0$   & $0.1$   & $0.2$   & $0.3$    & $\Delta$   \\
\midrule
$(2,2048)$ &  $34.99$ &	$35.50$ & $35.85$ & $36.12$  & $0.79$ \\
$(3,1024)$ &  $35.33$ &	$35.77$	& $35.94$ &	$35.83$  & $0.50$ \\
$(3,2048)$ &  $34.83$ &	$35.63$	& $36.08$ &	$\bm{36.22}$  & $0.52$ \\
$(4,1024)$ &  $35.45$ &	$35.72$	& $36.07$ &	$36.02$  & $0.88$ \\
$(4,2048)$ &  $35.06$ &	$35.61$	& $35.81$ &	$36.09$  & $0.63$ \\
\bottomrule
\end{tabular}
\label{tab:result_iwslt_deen}
\vspace{-1mm}
\end{table}

In summary, according to the exploration in Section~\ref{sec:hyper_explore} and Section~\ref{sec:proximal_init}, we empirically get the following conclusions: (1) A multi-branch attention layer is helpful to improve Transformer. We suggest to try $N_a=2$ or $N_a=3$ first. (2) A larger $d_h$ is helpful to bring better performance. Enlarging $d_h$ to consume the remaining parameters of reducing $d$ (of the standard Transformer) is a better choice. (3) The drop branch and proximal initialization are two key techniques to the success of MAT in NMT.

\subsection{Application to other NMT tasks}\label{sec:results_apply}

\noindent{\it Results of other IWSLT tasks}: We apply the discoveries to Spanish$\leftrightarrow$English and French$\leftrightarrow$English, and report the results in Table~\ref{tab:results_iwslt_esfren}.
For each setting, both the highest BLEU scores according to validation performance as well as the corresponding $\rho$'s are reported. We choose $1/512/1024$ as the default baseline, whose model size is the upper bound of our MAT models. We also implement two other baselines, one with smaller $d$ and the other with larger $d_h$. In baselines, the $\rho$ is fixed as zero.

Compared to the standard Transformer with one branch, various MAT with different model configurations outperform the baseline. The architecture $2/256/2048$ with $\rho=0.2$ generally obtains promising results. Compared with the default baseline $1/512/1024$, it achieves $1.74$, $1.90$, $1.31$, $1.25$ improvement on Es$\to$En, En$\to$Es, Fr$\to$En and En$\to$Fr. 

\begin{table}[!htbp]
\vspace{-2mm}
\centering
\caption{Results on IWSLT \{Es, Fr\}$\leftrightarrow$En. }
\begin{tabular}{c|ccc|ccccccc}
\toprule
$N_a$/$d$/$d_h$ & \#Params(M)  & Es$\to$En / $\rho$ & En$\to$Es / $\rho$ & \#Params(M)  & Es$\to$En / $\rho$ & En$\to$Es / $\rho$ \\
\midrule
1/512/1024 & $36.8$ & $40.37$ / $0.0$  & $38.56$ / $0.0$ & $36.8$ & $36.10$ / $0.0$ & $35.99$ / $0.0$ \\
1/256/1024 & $13.7$ & $40.60$ / $0.0$  & $39.42$ / $0.0$ & $13.7$ & $36.21$ / $0.0$ & $36.16$ / $0.0$ \\
1/256/2048 & $20.0$ & $40.18$ / $0.0$  & $38.78$ / $0.0$ & $20.0$ & $36.38$ / $0.0$ & $36.73$ / $0.0$ \\
\hline
2/256/1024 & $18.4$ & $41.02$ / $0.1$  & $39.56$ / $0.1$ & $18.4$ & $36.37$ / $0.2$ & $36.57$ / $0.2$ \\
2/256/2048 & $24.7$ & $42.11$ / $0.2$  & $40.46$ / $0.2$ & $24.7$ & $37.41$ / $0.2$ & $37.24$ / $0.2$ \\
3/256/1024 & $23.1$ & $41.39$ / $0.1$  & $40.08$ / $0.2$ & $23.2$ & $37.44$ / $0.2$ & $37.30$ / $0.1$ \\
3/256/2048 & $29.4$ & $41.79$ / $0.3$  & $40.40$ / $0.3$ & $29.5$ & $37.28$ / $0.2$ & $37.44$ / $0.2$ \\
\bottomrule
\end{tabular}
\label{tab:results_iwslt_esfren}
\vspace{-4mm}
\end{table}

\noindent{\it Results on larger datasets}: After obtaining results on small-scale datasets, we apply our discoveries to three larger datasets: WMT'14 En$\to$De translation, WMT'19 De$\to$Fr translation and WMT'19 En$\to$De translation.

The results of WMT'14 En$\to$De are shown in Table~\ref{tab:wmt14_en2de}. The standard baseline is $29.13$ BLEU score, and the model contains $209.8M$ parameters. We implement an MAT with $N_a=2,d_h=12288$ and another with $N_a=3,d_h=10240$. We can see that our MAT also works for the large-scale dataset. When $N_a=2$, we can obtain $29.90$ BLEU score. When increasing $N_a$ to $3$, we can get $29.85$, which is comparable with the the $N_a=2$ variant. Besides, for large-scale datasets, the drop branch technique is important too. $\rho=0.2$ works best of all settings. 

\begin{table}[!htbp]
\vspace{-2mm}
\centering
\caption{Results on WMT'14 En$\to$De translation. }
\begin{tabular}{lccccc}
\toprule
$N_a/d/d_h$ & \#Param (M) & $\rho$ & BLEU \\
\midrule
$1/1024/4096$ & $209.8$ & $0.0$ & $29.08$  \\
$1/512/10240$ & $161.7$ & $0.0$ & $28.86$ \\
$1/512/12288$ & $186.9$ & $0.0$ & $28.95$ \\
\midrule
$2/512/12288$ & $205.8$ & $0.1$ & $29.64$\\
$2/512/12288$ & $205.8$ & $0.2$ & $29.90$  \\
$2/512/12288$ & $205.8$ & $0.3$ & $29.06$  \\
$3/512/10240$ & $199.5$ & $0.1$ & $29.62$  \\
$3/512/10240$ & $199.5$ & $0.2$ & $29.85$  \\
$3/512/10240$ & $199.5$ & $0.3$ & $29.05$  \\
\bottomrule
\end{tabular}
\label{tab:wmt14_en2de}
\vspace{-1mm}
\end{table}

We summarize previous results on WMT'14 En$\to$De in Table~\ref{tab:prev_wmt14_en2de}. MAT can achieve comparable or slightly better results than carefully designed architectures like weighted Transformer and DynamicConv, and than the evolved Transformer discovered by neural architecture search.

The results of WMT'19 De$\to$Fr are shown in Table~\ref{tab:wmt19_de2fr}. Compared with standard big transformer model of architecture $1/1024/4096$, MAT with $\rho=0.1$ can improve the baseline by $0.58$ point.

\begin{table}[!htbp]
\vspace{-3mm}
\begin{minipage}{0.49\linewidth}
\vspace{-4mm}
\centering
\caption{BLEU scores of WMT'14 En$\to$De \\ in previous work.}
\begin{tabular}{lc}
\toprule
Algorithm & BLEU \\
\midrule
Weighted Transformer~\cite{ahmed2017weighted}& $28.9$ \\
Evolved Transformer~\cite{so2019evolved} & $29.8$ \\
DynamicConv~\cite{wu2019pay} & $29.7$ \\
\midrule
Our MAT & $29.9$\\
\bottomrule
\label{tab:prev_wmt14_en2de}
\end{tabular}
\end{minipage}
\hfill
\begin{minipage}{0.49\linewidth}
\centering
\caption{Results on WMT'19 De$\to$Fr translation.}
\begin{tabular}{lccccc}
\toprule
$N_a/d/d_h$ & \#Param (M) & $\rho$ & BLEU  \\
\midrule
$1/1024/4096$ & $223.6$ & $0.0$ & $34.07$  \\
$1/512/12288$ & $193.7$ & $0.0$ & $33.90$  \\
\midrule
$2/512/12288$ & $212.6$ & $0.0$ & $33.92$  \\
$2/512/12288$ & $212.6$ & $0.1$ & $34.65$ \\
$2/512/12288$ & $212.6$ & $0.2$ & $34.53$ \\
$2/512/12288$ & $212.6$ & $0.3$ & $34.49$ \\
\bottomrule
\label{tab:wmt19_de2fr}
\end{tabular}
\end{minipage}
\vspace{-3mm}
\end{table}

\noindent{\it Results with larger models}: Finally, we explore whether our proposed MAT works for larger models. We conduct experiments on WMT'19 En$\to$De translation task. The data is downloaded from WMT'19 website\footnote{The data is available at \url{http://www.statmt.org/wmt19/translation-task.html}. We concatenate \texttt{Europarl v9}, \texttt{Common Crawl corpus}, \texttt{News Commentary v14} and \texttt{Document-split Rapid} corpus. }. We change the number of encoder layers to $12$, set \texttt{attention-dropout} and \texttt{activation-dropout} as $0.1$, and keep the other settings the same as WMT'14 En$\to$De. To validate the effectiveness of MAT, we evaluate the trained models on three test sets: WMT'14 En$\to$De, WMT'18 En$\to$De and WMT'19 En$\to$De. We use \texttt{sacreBLEU} to evaluate the translation quality. Furthermore, to facilitate comparison with previous work, we also use \texttt{multi-bleu.perl} to calculate the BLEU score for WMT'14 En$\to$De.

The results are shown in Table~\ref{tab:wmt19_en2de}. Compared with the standard big transformer model of architecture $1/1024/4096$, in terms of \texttt{sacreBLEU}, MAT with $\rho=0.2$ can improve the baseline by $1.0$, $1.0$ and $1.1$ points on WMT14, WMT18 and WMT19 test sets respectively. Specially, on WMT'14 En$\to$De, in terms of \texttt{multi-bleu}, we achieve $30.8$ BLEU score, setting a new record on this work under the supervised setting.

\begin{table}[!htbp]
\vspace{-2mm}
\centering
\caption{Results of En$\to$De with larger models. For WMT'14, both \texttt{multi-bleu} (left) and \texttt{sacreBLEU} (right) are reported.}
\begin{tabular}{lcccccccc}
\toprule
$N_a/d/d_h$ & \#Param (M) & $\rho$ & WMT14  & WMT18 & WMT19  \\
\midrule
$1/1024/4096$ & $325.7$  & $0.0$ & $29.8$ / $29.2$ & $42.7$ & $39.3$ \\
$1/512/12288$ & $288.9$  & $0.0$ & $29.5$ / $29.0$ & $41.3$ & $37.4$ \\
\midrule
$2/512/12288$ & $314.1$  & $0.0$ & $30.0$ / $29.6$ & $42.5$ & $38.5$ \\
$2/512/12288$ & $314.1$  & $0.1$ & $29.9$ / $29.4$ & $43.1$ & $39.5$ \\
$2/512/12288$ & $314.1$  & $0.2$ & $30.8$ / $30.2$ & $43.7$ & $40.4$ \\
$2/512/12288$ & $314.1$  & $0.3$ & $30.1$ / $29.7$ & $43.8$ & $40.3$ \\
\bottomrule
\end{tabular}
\label{tab:wmt19_en2de}
\vspace{-2mm}
\end{table}

\section{Application to code generation}
We verify our proposed method on code generation, which is to map natural language sentences into code.

\noindent{\em Datasets} Following~\cite{wei2019code}, we conduct experiments on a Java dataset\footnote{The urls of the datasets and tools we used for code generation are summarized in Appendix C.\label{foot:codegen_script}}~\cite{ijcai2018-314} and a Python dataset~\cite{wan2018improving}. In the Java dataset, the numbers of training, validation and test sequences are $69708$, $8714$ and $8714$ respectively, and the corresponding numbers for Python are $55538$, $18505$ and $18502$. All samples are tokenized. We use the downloaded Java dataset without further processing and use Python standard AST module
to further process the python code. The source and target vocabulary sizes in natural language to Java code generation are $27k$ and $50k$, and those for natural language to Python code generation are $18k$ and $50k$. In this case, following~\cite{wei2019code}, we do not apply subword tokenization like BPE to the sequences.

\noindent{\em Model configuration} We set $N_a=4$, $d=256$ and $d_h=1024$ respectively. Both the encoder  and the decoder consist of three blocks. The MAT model contains about $37M$ parameters. For the Transformer baseline, we set $d=512$ and $d_h=1024$, with $55$M parameters.

\noindent{\em Evaluation} Following~\cite{wei2019code,ijcai2018-314}, we use sentence-level BLEU scores, which is the average BLEU score of all sequences to evaluate the generation quality. We choose the percentage of valid code (PoV) as another metric for evaluation, which is the percentage of code that can be parsed into an AST.

We report the results on code generation task in Table~\ref{tab:results_codegen}. It is obvious that Transformer-based models (standard Transformer and MAT) is significantly better than the LSTM-based dual model. Compared with standard Transformer, our MAT can get $4.2$ and $1.2$ BLEU score improvement in Java and Python datasets respectively. MAT reaches the best BLEU scores when $\rho=0.1$ on both on Java and Python datasets, which shows that the drop branch technique is important in code generation task. When $\rho=0.1$, in terms of PoV, our MAT can boost the Transformer baseline by $5.8\%$ and $6.6\%$ points. 

\begin{table}[!htbp]
\centering
\caption{Results on code generation.}
\begin{tabular}{lcccc}
\toprule
\multirow{2}{*}{Algorithm} & \multicolumn{2}{c}{Java} & \multicolumn{2}{c}{Python} \\
& BLEU & PoV & BLEU & PoV \\
\hline
Dual~\cite{wei2019code} & $17.17$ & $27.4\%$ & $12.09$ & $51.9\%$ \\
Transformer & $23.30$ & $83.1\%$ & $15.49$ & $73.3\%$ \\
\hline
Ours, $\rho=0.0$ & $26.97$ & $77.5\%$ & $16.35$ & $65.5\%$ \\
Ours, $\rho=0.1$ & $\bm{27.53}$ & $88.9\%$ & $\bm{16.66}$ & $79.9\%$ \\
Ours, $\rho=0.2$ & $26.06$ & $90.8\%$ & $15.63$ & $84.0\%$ \\
Ours, $\rho=0.3$ & $24.72$ & $90.2\%$ & $12.85$ & $75.3\%$ \\
\bottomrule
\end{tabular}
\label{tab:results_codegen}
\end{table}

\section{Application to natural language understanding}
We verify our proposed MAT on natural language understanding (NLU) tasks. Pre-training techniques have achieved state-of-the-art results on benchmark datasets/tasks like GLUE~\cite{wang2019glue}, RACE~\cite{lai2017large}, etc. To use pre-training, a masked language model is first pre-trained on large amount of unlabeled data. Then the obtained model is used to initialize weights for downstream tasks. We choose RoBERTa~\cite{liu2019roberta} as the backbone. For GLUE tasks, we focus more on accuracy. Therefore, we do not control the parameters subconsciously. We set $N_a=2$ for all experiments.

Following~\cite{fan2020reducing}, we conduct experiments on MNLI-m, MPRC, QNLI abd SST-2 tasks in GLUE benchmark~\cite{wang2019glue}. SST-2 is a single-sentence task, which is to check whether the input movie review is a positive one or negative. The remaining tasks are two-sentence tasks. In MNLI, given a premise sentence and a hypothesis sentence, the task is to predict whether the premise entails the hypothesis (entailment), contradicts the hypothesis (contradiction), or neither (neutral). In MRPC, the task is to check whether the two inputs are matched. In QNLI, the task is to determine whether the context sentence contains the answer to the question. 
We choose the 24-layer RoBERTa$_{\text{large}}$\footnote{Models from page \url{https://github.com/pytorch/fairseq/tree/master/examples/roberta}.} to initialize our MAT. We report the validation accuracy following~\cite{fan2020reducing}. 

\begin{table}[!htpb]
\centering
\caption{Validation results on various NLU tasks. Results of RoBERTa and RoBERTa + LayerDrop (briefly, ``+LayerDrop'') are from \cite{fan2020reducing}.} 
\begin{tabular}{lcccc}
\toprule
Algorithm & MNLI-m & MRPC & QNLI & SST-2 \\
\midrule
RoBERTa & $90.2$ & $90.9$ & $94.7$ & $96.4$ \\
+ LayerDrop & $90.1$ & $91.0$ & $94.7$ & $96.8$ \\
\midrule
Ours & $90.7$ & $91.9$ & $95.0$ & $97.0$ \\
\bottomrule
\end{tabular}
\label{tab:result_glue}
\end{table}

The results are reported in Table~\ref{tab:result_glue}. Compared with vanilla RoBERTa model, on the four tasks MNLI-m, MRPC, QNLI abd SST-2, we can improve them by $0.5$, $1.0$, $0.3$ and $0.6$ scores. Compared with the layer drop technique~\cite{fan2020reducing}, we also achieve promising improvement on these four tasks. The results demonstrate that our method not only works for sequence generation tasks, but also for text classification tasks. 

Drop branch also plays a central role on NLU tasks. We take the MRPC task as an example. The results are shown in Table~\ref{tab:mrpc_rho}. MRPC achieves the best result at $\rho=0.3$.

\begin{table}[!htbp]
\centering
\caption{MRPC validation accuracy w.r.t. $\rho$. }
\begin{tabular}{ccccccc}
\toprule
$\rho$ & $0.1$ & $0.2$ & $0.3$ & $0.4$ & $0.5$  \\
\midrule
accuracy & $90.9$ & $90.0$ & $91.9$ & $91.2$ & $91.7$ \\
\bottomrule
\end{tabular}
\label{tab:mrpc_rho}
\end{table}

\section{Exploring other variants}
In this section, we first discuss whether applying multi-branch architectures to FFN layers is helpful. Then we discuss a new variant of drop branch technique. Finally we verify the importance of the multi-branch architecture, where both concatenation and averaging operations are used. We mainly conduct experiments on IWSLT tasks to verify the variants. We conduct experiments on IWSLT'14 De$\to$En.

\subsection{Multi-branch FFN layer}
\noindent{\em Algorithm}: Denote the revised FFN layer as $\hffn_{N_f}$ with $N_f(\in\mathbb{Z}_+)$ branches. Mathematically, given an input $x$ from the previous layer,
\begin{equation}
\hffn_{N_f}(x)=x + \frac{1}{N_f}\sum_{i=1}^{N_f}\ffn(x;\omega_i)\frac{\mathbb{I}\{U_i\ge \rho\}}{1-\rho},
\end{equation}
where $\omega_i$ is the parameter of the $i$-th branch of the FFN layer, $x$ is the output from the previous layer. 

\noindent{\em Experiments}: We fix $d$ as $256$ and change $N_f$. We set the attention layers as both standard multi-head attention layers $(N_a=1)$ and multi-branch attention layers with $N_a=2$. We set $N_fd_h=2048$ to ensure the number of parameters unchanged with different number of branches. Results are reported in Table~\ref{tab:iwslt_deen_diff_ffn_branch}.

\begin{table}[!htbp]
	\centering
	\caption{Results on IWSLT'14 De$\to$En with multi-branch FFN layers. From left to right, the columns represent the network architecture (with the number of parameters included), BLEU scores with $\rho$ ranging from $0.0$ to $0.3$.}
	\begin{tabular}{lccccc}
		\toprule
		$N_a/N_f/d_h/$\#Param & $0.0$  & $0.1$ & $0.2$ & $0.3$ \\
		\midrule
		$1/1/2048/20.0 \text{M}$ & $34.66$ & $35.45$ & $32.75$ & $1.37$\\
		$1/2/1024/20.0 \text{M}$ & $34.85$ & $35.38$ & $35.30$ & $34.39$ \\
		$1/4/512/20.0 \text{M}$ & $34.51$ & $34.83$ & $34.71$ & $34.56$ \\
		$1/8/256/20.0 \text{M}$ & $34.35$ & $34.36$ & $34.38$ & $33.86$ \\
		\midrule
		$2/1/2048/24.7 \text{M}$ & $34.51$ &	$35.46$ & $35.59$ &	$35.33$ \\
		$2/2/1024/24.7 \text{M}$ & $34.39$ &	$35.18$ & $35.55$ &	$35.42$ \\
		$2/4/512/24.7 \text{M}$ & $33.86$ &	$34.90$ & $35.00$&	$35.07$ \\
		$2/8/256/24.7 \text{M}$ & $34.05$ &	$34.59$ &	$34.70$&	$34.55$ \\
		\bottomrule
	\end{tabular}
	\label{tab:iwslt_deen_diff_ffn_branch}
\end{table}

We can see that increasing the branches of FFN layer while decreasing the corresponding hidden dimension (i.e., $d_h$) will hurt the performance. For standard Transformer with $N_a=1$, as we increase $N_f$ from $1$ to $2$, $4$, $8$, the BLEU scores drop from $35.45$ to $35.38$, $34.83$ and $34.38$. For MAT with $N_a=2$, as $N_f$ grows from $1$ to $8$, the BLEU decreases from $35.59$ to $35.55$, $35.00$ and $34.70$. That is, constraint by the number of total parameters, we do not need to separate FFN layers into multiple branches.


\subsection{Variants of drop branch}

\noindent{\em Algorithm}: In drop branch, the attention heads within a branch are either dropped together or kept together. However, we are not sure whether dropping a complete branch is the best choice. Therefore, we design a more general way to leverage dropout. Mathematically, the $j$-th attention head of the $i$-th branch, denoted as $\beta^{i,j}(Q,K,V;\rho,\theta_{i,j})$, works as follows:
\begin{equation}
\begin{aligned}
\beta^{i,j}(Q,K,V;\rho,\theta_{i,j})=\attn(QW^{i,j}_Q, KW^{i,j}_K, VW^{i,j}_V)\frac{\mathbb{I}\{U_{i,j}\ge\rho\}}{1-\rho},
\end{aligned}
\label{eq:general_head}
\end{equation}
where superscripts $i$ and $j$ represent the branch id and head id in a multi-head attention layer respectively, $U^{i,j}\sim\texttt{unif}[0,1]$, $\theta_{i,j}=\{W^{i,j}_Q,W^{i,j}_K,W^{i,j}_V\}$. Based on Eqn.\eqref{eq:general_head}, we can define a more general multi-branch attentive architecture:
\begin{equation}
\begin{aligned}
&\frac{1}{N_a}\sum_{i=1}^{N_a}\concat\big(\beta^{i,1}(Q,K,V;\rho,\theta_{i,1}), \beta^{i,2}(Q,K,V;\rho,\theta_{i,2}),\cdots,\beta^{i,M}(Q,K,V;\rho,\theta_{i,M})
\big).
\end{aligned}
\label{eq:general_dropout}
\end{equation}
Keep in mind that Eqn.\eqref{eq:general_dropout} is associated with $U^{i,j}$ for any $i\in[N_a],j\in[M]$. We apply several constraints to $U^{i,j}$, which can lead to different ways to regularize the training:
\begin{enumerate}
\item For any $i\in[N_a]$, sample $U^i\sim\texttt{unif}[0,1]$ and set $U^{i,j}=U^i$ for any $j\in[M]$. This is the drop branch as we introduced in Section~\ref{sec:network_arch}.
\item Each $U^{i,j}$ is independently sampled.
\end{enumerate}

\noindent{\em Experiments}: We conduct experiments on IWSLT'14 De$\to$En and the results are reported in Table~\ref{tab:iwslt_deen_diff_dr_var}. Proximal initialization is leveraged. From left to right, each column represents architecture, number of parameters, BLEU scores with $\rho$ ranging from $0.0$ to $0.3$. The last column marked with $\Delta$ represents the improvement/decrease of the best results compared to those in Table~\ref{tab:result_iwslt_deen}.


\begin{table}[!htbp]
\vspace{-2mm}
\begin{minipage}{0.48\linewidth}
\centering
\small
\caption{Results on IWSLT'14 De$\to$En with randomly dropping heads technique. Embedding dimension $d$ is fixed as $256$.}
\begin{tabular}{p{1.0cm}p{0.57cm}p{0.57cm}p{0.57cm}p{0.57cm}p{0.57cm}}
\toprule
		$(N_a,d_h)$ & $0.0$  & $0.1$ & $0.2$ & $0.3$ & $\Delta$ \\
\midrule
		$(2,2048)$ & $34.99$ & $35.53$ & $35.95$ &	$\bm{35.96}$ & $-0.26$ \\
		$(3,1024)$ & $35.33$ & $35.67$ & $35.90$ &	$\bm{35.95}$ & $+0.01$\\
		$(3,2048)$ & $34.83$ & $35.45$ & $35.76$ &	$\bm{36.07}$ & $-0.15$\\
		$(4,1024)$ & $35.45$ & $35.72$ & $\bm{35.83}$ &	$35.77$ & $-0.24$\\
		$(4,2048)$ & $35.06$ & $35.79$ & $35.72$ &	$\bm{36.03}$ & $-0.06$\\
\bottomrule
\end{tabular}
\label{tab:iwslt_deen_diff_dr_var}
\end{minipage}
\hfill
\begin{minipage}{0.48\linewidth}
\vspace{-11mm}
\centering
\small
\caption{Results on IWSLT'14 De$\to$En \\ with different numbers of heads.}
\begin{tabular}{lcccc}
\toprule
$\rho$ & $0.1$ & $0.2$ & $0.3$\\
\midrule
$M=1$ & $35.01$ & $34.83$ & $34.52$\\
$M=2$ & $35.11$ & $35.45$ & $35.15$\\
$M=4$ & $35.19$ & $35.52$ & $35.11$\\
\bottomrule
\end{tabular}
\label{tab:iwslt_deen_m_heads_multi-branch}
\end{minipage}
\vspace{-2mm}
\end{table}

We can see that randomly dropping heads leads to slightly worse results than the drop branch technique. Take the network architecture with $N_a=3$, $d_h=1024$ as an example. The best BLEU score that drop branch technique achieves is $36.22$, and that obtained with randomly dropping heads is $36.07$. Similarly observations can be found from other settings in Table~\ref{tab:iwslt_deen_diff_dr_var}. Therefore, we suggest to use the drop branch technique, which requires minimal revision to the benchmark code.

\subsection{Ablation study on multi-branch attentions}
MAT introduces multi-branch attention, in addition to the multi-head attention with the concatenation manner.\footnote{The multi-head attention in standard Transformer is also one kind of multi-branch attention. For simplicity, we use multi-head attention to denote the original attention layer in standard Transformer, and multi-branch attention to denote the newly introduced multi-branch attention in the  multi-branch manner (both averaging and concatenation manner).} It is interesting to know whether the multi-head attention is still needed given the multi-branch attention. We conduct experiments with $N_a=2$, $d=256$, $d_h=1024$ and vary the number of heads $M$ in $\maattn_{N_a,M}$. Results are reported in Table~\ref{tab:iwslt_deen_m_heads_multi-branch}.



We can see that the multi-head attention is still helpful. When $M=1$, i.e., removing multiple heads and using a single head, it performs worse than multiple heads (e.g., $M=4$). Therefore, the multi-branch of them is the best choice.



\section{Conclusion and future work}
In this work, we proposed a simple yet effective variant of Transformer called multi-branch attentive Transformer and leveraged two techniques, drop branch and proximal initialization, to train the model. Rich experiments on neural machine translation, code generation and natural language understanding tasks have demonstrated the effectiveness of our model. 

For future work, we will apply our model to more sequence learning tasks. We will also combine our discoveries with neural architecture search, i.e., searching for better neural models for sequence learning in the search space with enriched multi-branch structures.




\appendix

\section*{Appendix}

\section{Exploring the Drop Branch in MAT}

We explore more values of drop branch rate $\rho$, and the results are reported in Table~\ref{tab:iwslt_deen_multi-branch_attn-more}.

\begin{table}[!htbp]
\centering
\caption{Results on IWSLT'14 De$\to$En with different architectures. From left to right, the columns represent the network architecture, number of parameters, BLEU scores with $\rho$ ranging from $0.2$ to $0.5$. Results of $\rho=0,0.1$ are in Table 1 of the main text.}
\begin{tabular}{lcccc}
\toprule
$N_a/d/d_h/\text{Param}$ &  $0.2$ & $0.3$ & $0.4$ & $0.5$ \\
\midrule
\multicolumn{5}{c}{Standard Transformer + Drop Branch} \\ 
\midrule
$1/512/1024/36.7M$ &    $22.43$ &   $0.78$ &    $1.64$ &    $0.00$ \\
$1/256/1024/13.7M$ &	$34.53$ &	$29.34$ &   $14.08$ &   $12.03$ \\
$1/256/2048/20.0M$ &	$32.75$ &	$1.37$ &    $1.48$ &    $9.05$ \\
$1/256/3072/26.3M$ & 	$34.64$ &	$25.05$ &   $1.31$ &    $7.84$ \\
\midrule
\multicolumn{5}{c}{MAT + Drop Branch} \\
\midrule
$2/256/1024/18.4M$ &    $35.52$ & $35.11$ &    $34.46$ &   $33.50$ \\
$2/256/2048/24.7M$ &    $35.59$ & $35.33$ &    $34.95$ &   $28.15$ \\
$3/256/1024/23.1M$ &    $35.39$&	$35.44$ &   $34.85$ &   $34.10$ \\
$3/256/2048/29.4M$ &    $35.40$ &	$\bm{35.70}$ &  $35.01$ &   $31.68$ \\
$4/256/1024/27.9M$ & 	$35.08$ &	$35.14$ &   $35.01$ &   $34.38$ \\
$4/256/2048/34.2M$ & 	$35.08$ &	$35.46$ &   $34.21$ &   $0.89$ \\
\bottomrule
\end{tabular}
\label{tab:iwslt_deen_multi-branch_attn-more}
\end{table}

From Table~\ref{tab:iwslt_deen_multi-branch_attn-more}, we can see that increasing $\rho$ to $0.4$ and $0.5$ will hurt the performance of MAT.

\begin{table}[!htbp]
\centering
\caption{Results of using proximal initialization. Columns from left to right represent the network architecture, BLEU scores with drop branch ratio $\rho$ from $0.2$ to $0.5$, and the increment compared to the best results without proximal initialization.}
\begin{tabular}{cccccc}
\toprule
$(N_a,d_h)$&  $0.2$   & $0.3$   & $0.4$   & $0.5$   & $\Delta$   \\
\midrule
$(2,2048)$ &  $35.85$ & $36.12$ & $35.62$ & $35.03$ & $0.79$ \\
$(3,1024)$ &  $35.94$ &	$35.83$ & $35.76$ & $35.15$ & $0.50$ \\
$(3,2048)$ &  $36.08$ &	$\bm{36.22}$ & $35.87$ & $35.58$ & $0.52$ \\
$(4,1024)$ &  $36.07$ &	$36.02$ & $35.78$ & $35.42$ & $0.88$ \\
$(4,2048)$ &  $35.81$ &	$36.09$ & $35.80$ & $35.57$ & $0.63$ \\
\bottomrule
\end{tabular}
\label{tab:result_iwslt_deen-more}
\end{table}

On IWSLT'14 De$\to$En, with proximal initialization, we also explore increasing $\rho$ to $0.4$ and $0.5$. The results are in Table~\ref{tab:result_iwslt_deen-more}. Still, setting $\rho$ larger than $0.4$ will hurt the performance.

\section{Exploration on Other IWSLT Tasks}

\subsection{Results of MAT}

We apply the settings in Table~$1$ of the main paper to all four other IWSLT tasks. The results without proximal initialization are reported from Table~\ref{tab:iwslt_esen_multi-branch_attn} to Table~\ref{tab:iwslt_enfr_multi-branch_attn}. The results with promixal initialization are shown from Table~\ref{tab:result_iwslt_esen} to Table~\ref{tab:result_iwslt_enfr}.

\begin{table}[!htbp]
\centering
\small
\caption{Results on IWSLT Es$\to$En with different architectures. From left to right, the columns represent the network architecture, number of parameters, BLEU scores with $\rho$ ranging from $0.0$ to $0.3$.}
\begin{tabular}{lcccccc}
\toprule
$N_a/d/d_h/\text{Param}$ &  $0.0$ & $0.1$ & $0.2$ & $0.3$  \\
\midrule
\multicolumn{5}{c}{Standard Transformer + Drop Branch} \\ 
\midrule    
$1/512/1024/36.7M$ &    $39.77^\triangle$ & $2.19$ &    $0.00$ &   $0.00$ \\
$1/256/1024/13.7M$ &	$40.60$ &	$40.94$ &	$39.46$ &	$26.04$  \\
$1/256/2048/20.0M$ &	$40.18$ &	$41.30$ &   $39.16$ &	$0.54$  \\
\midrule
\multicolumn{5}{c}{MAT + Drop Branch} \\
\midrule
$2/256/1024/18.4M$ &    $40.30$ & $41.02$ & $40.89$ & $40.45$ \\
$2/256/2048/24.7M$ &    $40.28$ & $41.20$ & $\bm{41.96}$ & $41.06$  \\
$3/256/1024/23.1M$ &    $40.42$ & $41.04$ & $40.99$ & $40.59$  \\
$3/256/2048/29.4M$ &    $39.74$ & $40.72$ & $41.63$ & $40.80$ \\
$4/256/1024/27.9M$ & 	$40.17$ & $40.99$ & $40.81$ & $40.67$ \\
$4/256/2048/34.2M$ & 	$39.94$ & $40.69$ & $41.29$ & $40.01$ \\
\bottomrule
\end{tabular}
\label{tab:iwslt_esen_multi-branch_attn}
\end{table}

\begin{table}[!htbp]
\centering
\small
\caption{Results on IWSLT En$\to$Es with different architectures. From left to right, the columns represent the network architecture, number of parameters, BLEU scores with $\rho$ ranging from $0.0$ to $0.3$.}
\begin{tabular}{lcccccc}
\toprule
$N_a/d/d_h/\text{Param}$ &  $0.0$ & $0.1$ & $0.2$ & $0.3$  \\
\midrule
\multicolumn{5}{c}{Standard Transformer + Drop Branch} \\ 
\midrule
$1/512/1024/36.7M$ &    $38.56^\triangle$ & $1.67$ &    $0.01$ &   $0.00$ \\
$1/256/1024/13.7M$ &	$39.51$ &	$39.58$ &	$38.42$ &	$23.50$  \\
$1/256/2048/20.0M$ &	$38.78$ &	$39.84$ &   $38.22$ &	$0.90$  \\   
\midrule
\multicolumn{5}{c}{MAT + Drop Branch} \\
\midrule
$2/256/1024/18.4M$ &    $38.81$ & $39.56$ & $38.98$ & $39.08$ \\
$2/256/2048/24.7M$ &    $38.77$ & $39.33$ & $39.86$ & $39.37$  \\
$3/256/1024/23.1M$ &    $38.31$ & $39.12$ & $39.77$ & $39.29$  \\
$3/256/2048/29.4M$ &    $38.22$ & $39.89$ & $39.57$ & $\bm{39.90}$ \\
$4/256/1024/27.9M$ & 	$38.37$ & $39.16$ & $39.55$ & $39.49$ \\
$4/256/2048/34.2M$ & 	$38.40$ & $39.05$ & $39.24$ & $39.35$ \\
\bottomrule
\end{tabular}
\label{tab:iwslt_enes_multi-branch_attn}
\end{table}

\begin{table}[!htbp]
\centering
\small
\caption{Results on IWSLT Fr$\to$En with different architectures. From left to right, the columns represent the network architecture, number of parameters, BLEU scores with $\rho$ ranging from $0.0$ to $0.3$.}
\begin{tabular}{lcccccc}
\toprule
$N_a/d/d_h/\text{Param}$ &  $0.0$ & $0.1$ & $0.2$ & $0.3$  \\
\midrule
\multicolumn{5}{c}{Standard Transformer + Drop Branch} \\ 
\midrule
$1/512/1024/36.7M$ &    $36.60^\triangle$ & $2.09$ &    $0.01$ &   $0.01$ \\
$1/256/1024/13.7M$ &	$36.21$ &	$36.17$ &	$35.15$ &	$3.99$  \\
$1/256/2048/20.0M$ &	$36.38$ &	$36.42$ &   $35.11$ &	$14.66$  \\
\midrule
\multicolumn{5}{c}{MAT + Drop Branch} \\
\midrule
$2/256/1024/18.4M$ &    $35.63$ & $36.28$ & $36.37$ & $35.61$ \\
$2/256/2048/24.7M$ &    $35.90$ & $35.83$ & $36.34$ & $36.08$  \\
$3/256/1024/23.1M$ &    $35.72$ & $35.71$ & $36.11$ & $35.59$  \\
$3/256/2048/29.4M$ &    $35.55$ & $36.30$ & $36.22$ & $36.27$ \\
$4/256/1024/27.9M$ & 	$35.52$ & $36.30$ & $36.06$ & $36.12$ \\
$4/256/2048/34.2M$ & 	$34.98$ & $\bm{36.67}$ & $36.24$ & $35.97$ \\
\bottomrule
\end{tabular}
\label{tab:iwslt_fren_multi-branch_attn}
\end{table}

\begin{table}[!htbp]
\centering
\small
\caption{Results on IWSLT En$\to$Fr with different architectures. From left to right, the columns represent the network architecture, number of parameters, BLEU scores with $\rho$ ranging from $0.0$ to $0.3$.}
\begin{tabular}{lcccccc}
\toprule
$N_a/d/d_h/\text{Param}$ &  $0.0$ & $0.1$ & $0.2$ & $0.3$  \\
\midrule
\multicolumn{5}{c}{Standard Transformer + Drop Branch} \\ 
\midrule
$1/512/1024/36.7M$ &    $36.32^\triangle$ & $36.07$ &    $0.02$ &   $0.05$ \\
$1/256/1024/13.7M$ &	$36.16$ &	$36.43$ &	$35.61$ &	$0.79$  \\
$1/256/2048/20.0M$ &	$36.73$ &	$36.80$ &   $35.34$ &	$1.59$  \\
\midrule
\multicolumn{5}{c}{MAT + Drop Branch} \\
\midrule
$2/256/1024/18.4M$ &    $36.41$ & $36.48$ & $36.57$ & $36.34$ \\
$2/256/2048/24.7M$ &    $35.95$ & $\bm{37.20}$ & $36.90$ & $36.44$  \\
$3/256/1024/23.1M$ &    $35.79$ & $36.52$ & $36.56$ & $36.34$  \\
$3/256/2048/29.4M$ &    $35.88$ & $36.88$ & $36.88$ & $36.86$ \\
$4/256/1024/27.9M$ & 	$35.77$ & $36.68$ & $37.09$ & $36.50$ \\
$4/256/2048/34.2M$ & 	$35.23$ & $36.65$ & $36.97$ & $36.66$ \\
\bottomrule
\end{tabular}
\label{tab:iwslt_enfr_multi-branch_attn}
\end{table}

\begin{table}[!htbp]
\centering
\caption{Results of using proximal initialization on IWSLT Es$\to$En. Columns from left to right represent the network architecture, BLEU scores with drop branch ratio $\rho$ from $0.0$ to $0.3$, and the increment compared to the best results without proximal initialization.}
\begin{tabular}{cccccccc}
\toprule
$(N_a,d_h)$&  $0.0$   & $0.1$   & $0.2$   & $0.3$    & $\Delta$   \\
\midrule
$(2,2048)$ &  $40.81$ &	$41.14$ & $\bm{42.11}$ & $41.68$  & $0.15$ \\
$(3,1024)$ &  $40.95$ &	$41.39$	& $41.22$ &	$41.36$  & $0.35$ \\
$(3,2048)$ &  $40.96$ &	$41.10$	& $41.60$ &	$41.79$  & $0.16$ \\
$(4,1024)$ &  $41.45$ & $41.67$	& $41.67$ & $41.26$  & $0.68$ \\
$(4,2048)$ &  $40.85$ &	$41.44$	& $41.26$ &	$42.06$  & $0.77$ \\
\bottomrule
\end{tabular}
\label{tab:result_iwslt_esen}
\end{table}

\begin{table}[!htbp]
\centering
\caption{Results of using proximal initialization on IWSLT En$\to$Es. Columns from left to right represent the network architecture, BLEU scores with drop branch ratio $\rho$ from $0.0$ to $0.3$, and the increment compared to the best results without proximal initialization.}
\begin{tabular}{cccccccc}
\toprule
$(N_a,d_h)$&  $0.0$   & $0.1$   & $0.2$   & $0.3$    & $\Delta$   \\
\midrule
$(2,2048)$ &  $39.47$ &	$39.78$ & $40.46$ & $40.34$  & $0.60$ \\
$(3,1024)$ &  $39.47$ &	$40.06$	& $40.08$ &	$39.84$  & $0.31$ \\
$(3,2048)$ &  $39.34$ &	$39.61$	& $40.08$ &	$\bm{40.40}$  & $0.50$ \\
$(4,1024)$ &  $39.36$ & $39.99$	& $39.81$ & $40.00$  & $0.45$ \\
$(4,2048)$ &  $39.75$ &	$40.16$	& $40.12$ &	$39.91$  & $0.81$ \\
\bottomrule
\end{tabular}
\label{tab:result_iwslt_enes}
\end{table}

\begin{table}[!htbp]
\centering
\caption{Results of using proximal initialization on IWSLT Fr$\to$En. Columns from left to right represent the network architecture, BLEU scores with drop branch ratio $\rho$ from $0.0$ to $0.3$, and the increment compared to the best results without proximal initialization.}
\begin{tabular}{cccccccc}
\toprule
$(N_a,d_h)$&  $0.0$   & $0.1$   & $0.2$   & $0.3$    & $\Delta$   \\
\midrule
$(2,2048)$ &  $36.14$ &	$36.76$ & $37.41$ & $36.82$  & $0.77$ \\
$(3,1024)$ &  $36.73$ &	$37.03$	& $\bm{37.44}$ &	$37.10$  & $1.33$ \\
$(3,2048)$ &  $36.32$ &	$36.92$	& $37.28$ &	$37.25$  & $0.98$ \\
$(4,1024)$ &  $36.81$ & $37.14$	& $37.10$ & $36.88$  & $0.73$ \\
$(4,2048)$ &  $36.48$ &	$36.90$	& $37.40$ &	$37.29$  & $0.73$ \\
\bottomrule
\end{tabular}
\label{tab:result_iwslt_fren}
\end{table}

\begin{table}[!htbp]
\centering
\caption{Results of using proximal initialization on IWSLT En$\to$Fr. Columns from left to right represent the network architecture, BLEU scores with drop branch ratio $\rho$ from $0.0$ to $0.3$, and the increment compared to the best results without proximal initialization.}
\begin{tabular}{cccccccc}
\toprule
$(N_a,d_h)$&  $0.0$   & $0.1$   & $0.2$   & $0.3$    & $\Delta$   \\
\midrule
$(2,2048)$ &  $36.73$ &	$37.14$ & $37.24$ & $37.09$  & $0.04$ \\
$(3,1024)$ &  $36.71$ &	$37.30$	& $37.04$ &	$36.92$  & $0.74$ \\
$(3,2048)$ &  $36.51$ &	$37.39$	& $\bm{37.44}$ &	$37.43$  & $0.56$ \\
$(4,1024)$ &  $36.67$ & $36.77$	& $37.24$ & $37.15$  & $0.15$ \\
$(4,2048)$ &  $36.70$ &	$37.19$	& $37.18$ &	$37.32$  & $0.35$ \\
\bottomrule
\end{tabular}
\label{tab:result_iwslt_enfr}
\end{table}

\subsection{Variants of Drop Branch}
We explore the variant of drop branch described in Section 7.2 on the other four IWSLT tasks. We apply all settings in Table~11 of the main paper to the remaining language pairs. Results are reported from Table~\ref{tab:iwslt_esen_diff_dr_var} to Table~\ref{tab:iwslt_enfr_diff_dr_var}. Generally, the two types of drop branch achieve similar results.

\begin{table}[!htbp]
\centering
\caption{Results on IWSLT Es$\to$En with randomly dropping heads technique. Embedding dimension $d$ is fixed as $256$.}
\begin{tabular}{lccccc}
\toprule
		$(N_a,d_h)$ & $0.0$  & $0.1$ & $0.2$ & $0.3$ & $\Delta$ \\
\midrule
		$(2,2048)$ & $40.81$ & $41.61$ & $41.54$ &	$\bm{41.68}$ & $-0.43$ \\
		$(3,1024)$ & $40.95$ & $\bm{41.63}$ & $41.25$ &	$41.15$ & $+0.24$ \\
		$(3,2048)$ & $40.96$ & $41.30$ & $41.39$ &	$\bm{41.86}$ & $+0.07$ \\
		$(4,1024)$ & $\bm{41.45}$ & $41.41$ & $41.04$ &	$41.29$ & $-0.22$ \\
		$(4,2048)$ & $40.85$ & $41.30$ & $41.48$ &	$\bm{41.70}$ & $-0.36$ \\
\bottomrule
\end{tabular}
\label{tab:iwslt_esen_diff_dr_var}
\end{table}

\begin{table}[!htbp]
\centering
\caption{Results on IWSLT En$\to$Es with randomly dropping heads technique. Embedding dimension $d$ is fixed as $256$.}
\begin{tabular}{lccccc}
\toprule
		$(N_a,d_h)$ & $0.0$  & $0.1$ & $0.2$ & $0.3$ & $\Delta$ \\
\midrule
		$(2,2048)$ & $39.47$ & $39.85$ & $39.98$ &	$\bm{40.39}$ & $-0.07$ \\
		$(3,1024)$ & $39.47$ & $\bm{39.99}$ & $39.96$ &	$39.41$ & $-0.09$ \\
		$(3,2048)$ & $39.34$ & $39.47$ & $\bm{40.22}$ &	$39.67$ & $-0.18$ \\
		$(4,1024)$ & $39.36$ & $39.71$ & $39.59$ &	$\bm{39.89}$ & $-0.11$ \\
		$(4,2048)$ & $39.75$ & $39.73$ & $39.94$ &	$\bm{40.36}$ & $+0.20$ \\
\bottomrule
\end{tabular}
\label{tab:iwslt_enes_diff_dr_var}
\end{table}

\begin{table}[!htbp]
\centering
\caption{Results on IWSLT Fr$\to$En with randomly dropping heads technique. Embedding dimension $d$ is fixed as $256$.}
\begin{tabular}{lccccc}
\toprule
		$(N_a,d_h)$ & $0.0$  & $0.1$ & $0.2$ & $0.3$ & $\Delta$ \\
\midrule
		$(2,2048)$ & $36.14$ & $\bm{37.10}$ & $36.85$ &	$36.71$ & $-0.31$ \\
		$(3,1024)$ & $36.73$ & $36.78$ & $\bm{36.83}$ &	$36.20$ & $-0.61$ \\
		$(3,2048)$ & $36.32$ & $36.73$ & $36.92$ &	$\bm{37.08}$ & $-0.20$ \\
		$(4,1024)$ & $36.81$ & $36.29$ & $\bm{36.82}$ &	$36.49$ & $-0.32$ \\
		$(4,2048)$ & $36.48$ & $36.39$ & $36.89$ &	$\bm{36.99}$ & $-0.41$ \\
\bottomrule
\end{tabular}
\label{tab:iwslt_fren_diff_dr_var}
\end{table}

\begin{table}[!htbp]
\centering
\caption{Results on IWSLT En$\to$Fr with randomly dropping heads technique. Embedding dimension $d$ is fixed as $256$.}
\begin{tabular}{lccccc}
\toprule
		$(N_a,d_h)$ & $0.0$  & $0.1$ & $0.2$ & $0.3$ & $\Delta$ \\
\midrule
		$(2,2048)$ & $36.73$ & $37.18$ & $\bm{37.48}$ &	$37.15$ & $+0.24$ \\
		$(3,1024)$ & $36.71$ & $37.14$ & $\bm{37.48}$ &	$36.86$ & $+0.18$ \\
		$(3,2048)$ & $36.51$ & $36.93$ & $\bm{37.17}$ &	$36.97$ & $-0.27$ \\
		$(4,1024)$ & $36.67$ & $36.59$ & $\bm{37.02}$ &	$36.99$ & $-0.22$ \\
		$(4,2048)$ & $36.70$ & $36.81$ & $\bm{37.80}$ &	$37.37$ & $+0.48$ \\
\bottomrule
\end{tabular}
\label{tab:iwslt_enfr_diff_dr_var}
\end{table}

\section{Scripts}
In this section, we summarize the scripts we used in our paper.

\noindent(1) \texttt{multi-bleu.perl}: \url{https://github.com/moses-smt/mosesdecoder/blob/master/scripts/generic/multi-bleu.perl}

\noindent(2) \texttt{sacreBLEU}: \url{https://github.com/mjpost/sacreBLEU}

\noindent(3) The script to preprocess the IWSLT data: \url{https://github.com/pytorch/fairseq/blob/master/examples/translation/prepare-iwslt14.sh}\label{foot:fb_script}

\noindent(4) Path to Java dataset: \url{https://github.com/xing-hu/TL-CodeSum}

\noindent(5) Path to Python dataset: \url{https://github.com/wanyao1992/code_summarization_public}

\noindent(6) Python standard AST module to process the Python code: \url{https://docs.python.org/3/library/ast.html}

\bibliography{mybib}
\bibliographystyle{plain}

\end{document}